\documentclass[11pt,a4paper]{article}
\usepackage[hyperref]{acl2019}
\usepackage{times}
\usepackage{latexsym}

\usepackage{url}
\usepackage{graphicx}  
\usepackage{amsfonts}
\usepackage{CJK}
\usepackage{amsmath}
\usepackage{bm}
\usepackage{mathrsfs}
\usepackage{float}
\usepackage{xcolor,colortbl}
\usepackage{listings}

\usepackage{subfigure}
\usepackage{booktabs,multirow}
\usepackage{bigstrut,bigdelim}
\usepackage{paralist}
\usepackage{bbm}

\aclfinalcopy 
\def\aclpaperid{1113} 

\title{Learning a Matching Model with Co-teaching for Multi-turn Response Selection in Retrieval-based Dialogue Systems}

\author{Jiazhan Feng$^{1}$\thanks{\ \ Equal Contribution.}\ \ , Chongyang Tao$^{1}$\footnotemark[1]\ \ , Wei Wu$^2$, Yansong Feng$^{1}$, 
\\\textbf{Dongyan Zhao}$^{1,3}$ \and \textbf{Rui Yan}$^{1,3}$\thanks{\ \ Corresponding author: Rui Yan (ruiyan@pku.edu.cn).} \\
$^1$Institute of Computer Science and Technology, Peking University, Beijing, China \\
$^2$Microsoft Corporation, Beijing, China\\
$^3$Center for Data Science, Peking University, Beijing, China \\
{\tt $^{1}$fengjiazhan@foxmail.com \quad $^{2}$wuwei@microsoft.com} \\
{\tt $^{1,3}$\{chongyangtao,fengyansong,zhaody,ruiyan\}@pku.edu.cn} \\
}

\date{}
\usepackage[linesnumbered, ruled]{algorithm2e} 
\usepackage{algpseudocode}

\begin{document}

\maketitle

\begin{abstract}
We study learning of a matching model for response selection in retrieval-based dialogue systems. The problem is equally important with designing the architecture of a model, but is less explored in existing literature. To learn a robust matching model from noisy training data, we propose a general co-teaching framework with three specific teaching strategies that cover both teaching with loss functions and teaching with data curriculum. Under the framework, we simultaneously learn two matching models with independent training sets. In each iteration, one model transfers the knowledge learned from its training set to the other model, and at the same time receives the guide from the other model on how to overcome noise in training. Through being both a teacher and a student, the two models learn from each other and get improved together.  Evaluation results on two public data sets indicate that the proposed learning approach can generally and significantly improve the performance of existing matching models.
\end{abstract}

\section{Introduction} 
Human-machine conversation is a long-standing goal of artificial intelligence. Recently, building a dialogue system for open domain human-machine conversation is attracting more and more attention due to both availability of large-scale human conversation data and powerful models learned with neural networks. Existing methods are either retrieval-based or generation-based. Retrieval-based methods reply to a human input by selecting a proper response from a pre-built index \cite{ji2014information,zhou2018multi,yan2018coupled}, while generation-based methods synthesize a response with a natural language model \cite{shangL2015neural,serban2016multiresolution}. In this work, we study the problem of response selection for retrieval-based dialogue systems, since retrieval-based systems are often superior to their generation-based counterparts on response fluency and diversity, are easy to evaluate, and have powered some real products such as the social bot XiaoIce from Microsoft \cite{shum2018eliza}, and the E-commerce assistant AliMe Assist from Alibaba Group \cite{li2017alime}. 

A key problem in response selection is how to measure the matching degree between a conversation context (a message with several turns of conversation history) and a response candidate. Existing studies have paid tremendous effort to build a matching model with neural architectures \cite{lowe2015ubuntu,zhou2016multi,wu2017sequential,zhou2018multi}, and advanced models such as the deep attention matching network (DAM) \cite{zhou2018multi} have achieved impressive performance on benchmarks. In contrary to the progress on model architectures, there is little exploration on learning approaches of the models. On the one hand, neural matching models are becoming more and more complicated; on the other hand, all models are simply learned by distinguishing human responses from some automatically constructed negative response candidates (e.g., by random sampling). Although this heuristic approach can avoid expensive and exhausting human labeling, it suffers from noise in training data, as many negative examples are actually false negatives\footnote{Responses sampled from other contexts may also be proper candidates for a given context.}. As a result, when evaluating a well-trained model using human judgment, one can often observe a significant gap between training and test, as will be seen in our experiments. 

In this paper, instead of configuring new architectures, we investigate how to effectively learn existing matching models from noisy training data, given that human labeling is infeasible in practice. We propose learning a matching model under a general co-teaching framework. The framework maintains two peer models on two i.i.d. training sets, and lets the two models teach each other during learning. One model transfers  knowledge learned from its training set to its peer model to help it combat with noise in training, and at the same time gets updated under the guide of its peer model. Through playing both a role of a teacher and a role of a student, the two peer models evolve together. Under the  framework, we consider three teaching strategies including teaching with dynamic margins, teaching with dynamic instance weighting, and teaching with dynamic data curriculum. The first two strategies let the two peer models mutually ``label'' their training examples, and transfer the soft labels from one model to the other through loss functions;  while in the last strategy, the two peer models directly select training examples for each other.

To examine if the proposed learning approach can generally bridge the gap between training and test, we select sequential matching network (SMN) \cite{wu2017sequential} and DAM as representative matching models, and conduct experiments on two public data sets with human judged test examples. The first data set is the Douban Conversation benchmark published in \newcite{wu2017sequential}, and the second one is the E-commerce Dialogue Corpus published in \newcite{coling2018dua} where we recruit human annotators to judge the appropriateness of response candidates regarding to their contexts on the entire test set\footnote{We have released labeled test data of E-commerce Dialogue Corpus at \url{https://drive.google.com/open?id=1HMDHRU8kbbWTsPVr6lKU_-Z2Jt-n-dys}.}. Evaluation results indicate that co-teaching with the three strategies can consistently improve the performance of both matching models over all metrics on both data sets with significant margins.  On the Douban data, the most effective strategy is teaching with dynamic margins that brings $2.8$\% absolute improvement to SMN and $2.5$\% absolute improvement to DAM on P@1; while on the E-commerce data, the best strategy is teaching with dynamic data curriculum that brings $2.4$\% absolute improvement to SMN and $3.2$\% absolute improvement to DAM on P@1. Through further analysis, we also unveil how the peer models get evolved together in learning and how the choice of peer models affects the performance of learning.

Our contributions in the paper are four-folds: (1) proposal of learning matching models for response selection with a general co-teaching framework; (2) proposal of two new teaching strategies as special cases of the framework; and (3) empirical verification of the effectiveness of the proposed learning approach on two public data sets.
\section{Problem Formalization}
Given a data set $\mathcal {D} = \{(y_i,c_i,r_i)\}_{i=1}^N$ where $c_i$ represents a conversation context, $r_i$ is a response candidate, and $y_i\in \{0,1\}$ denotes a label with $y_i=1$ indicating $r_i$ a proper response for $c_i$ and otherwise $y_i=0$, the goal of the task of response selection is to learn a matching model $s(\cdot,\cdot)$ from $\mathcal{D}$. For any context-response pair $(c,r)$, $s(c,r)$ gives a score that reflects the matching degree between $c$ and $r$, and thus allows one to rank a set of response candidates according to the scores for response selection.

To obtain a matching model $s(\cdot,\cdot)$, one needs to deal with two problems: (1) how to define $s(\cdot,\cdot)$; and (2) how to learn $s(\cdot,\cdot)$. Existing studies concentrate on Problem (1) by defining $s(\cdot,\cdot)$ with sophisticated neural architectures \cite{wu2017sequential,zhou2018multi}, and leave Problem (2) in a simple default setting where $s(\cdot,\cdot)$ is optimized with $\mathcal{D}$ using a loss function $L$ usually defined by cross entropy.  Ideally, when $\mathcal{D}$ is large enough and has good enough quality, a carefully designed $s(\cdot,\cdot)$ learned using the existing paradigm should be able to well capture the semantics in dialogues. The fact is that since large-scale human labeling is infeasible, $\mathcal{D}$ is established under simple heuristics where negative response candidates are automatically constructed (e.g., by random sampling) with a lot of noise. As a result, advanced matching models only have sub-optimal performance in practice. The gap between ideal and reality motivates us to pursue a better learning approach, as will be presented in the next section.

\section{Learning a Matching Model through Co-teaching}
In this section, we present co-teaching, a new framework for learning a matching model. We first give a general description of the framework, and then elaborate three teaching strategies as special cases of the framework.

\begin{figure}[ht!]
    \centering
    \includegraphics[width=0.47\textwidth]{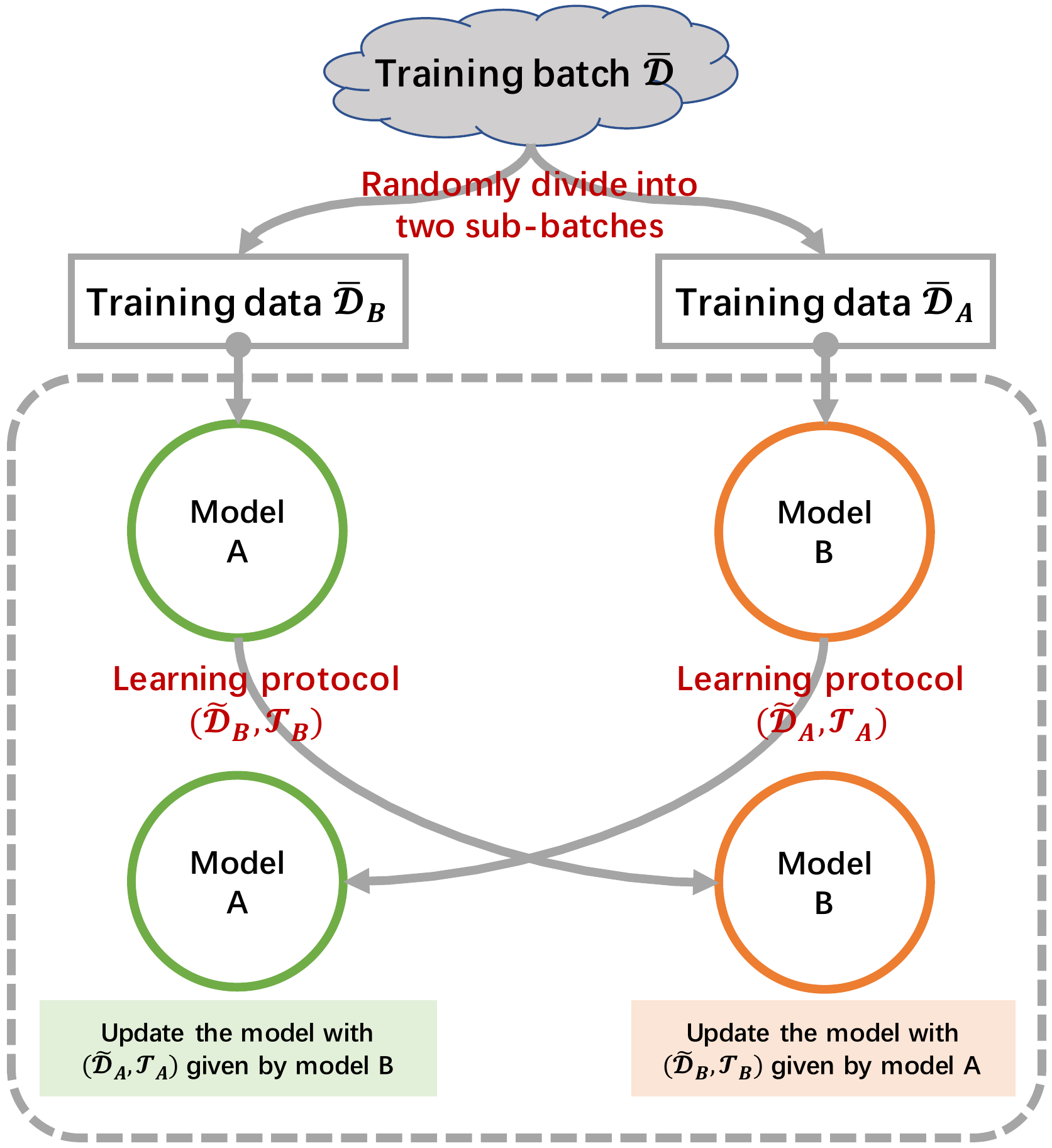}
    \caption{Co-teaching framework.}
    \label{fig:framework overview}
\end{figure}

\subsection{Co-teaching Framework}
The idea of co-teaching is to maintain two peer models and let them learn from each other by simultaneously acting as a teacher and a student. 

Figure \ref{fig:framework overview} gives an overview of the co-teaching framework. The learning program starts from two pre-trained peer models A and B.  In each iteration, a batch of training data is equally divided into two sub-batches without overlap as $\Bar{\mathcal{D}}_A$ and $\Bar{\mathcal{D}}_B$ for B and A respectively.  A and B then examine their sub-batches and output learning protocols $(\Tilde{\mathcal{D}}_B, \mathcal{J}_B)$ and  $(\Tilde{\mathcal{D}}_A, \mathcal{J}_A)$ for their peers, where $\Tilde{\mathcal{D}}_B$ and $\Tilde{\mathcal{D}}_A$ are training data and $\mathcal{J}_B$ and $\mathcal{J}_A$ are loss functions. After that, A and B get updated according to $(\Tilde{\mathcal{D}}_A, \mathcal{J}_A)$ and $(\Tilde{\mathcal{D}}_B, \mathcal{J}_B)$ respectively, and the learning program moves to the next iteration. Algorithm 1 describes the pseudo code of co-teaching. 

The rationale behind the co-teaching framework is that the peer models can gradually obtain different abilities from the different training data as the learning process goes on, even when the two models share the same architecture and the same initial configuration, and thus, they can acquire different knowledge from their training data and transfer the knowledge to their peers to make them robust over the noise in the data.  This resembles two peer students who learn from different but related materials. Through knowledge exchange, one can inspire the other to get new insights from his or her material, and thus the two students get improved together. Advantages of the framework reside in various aspects: first, the peer models have their own ``judgment'' regarding to the quality of the same training example. Thus, one model may guide the other how to pick high quality training examples and circumvent noise; second, since the peer models are optimized with different training sub-batches, knowledge from one sub-batch could be supplementary to the other through exchange of learning protocols; third, the two peer models may have different decision boundaries, and thus are good at recognizing different  patterns in data. This may allow one model to help the other rectify errors in learning.  

To instantiate the co-teaching framework, one needs to specify initialization of the peer models and teaching strategies that can form the learning protocols. In this work, to simplify the learning program of co-teaching, we assume that model A and model B are initialized by the same matching model pre-trained with the entire training data.  We focus on design of teaching strategies, as will be elaborated in the next section. 

\begin{algorithm*}[t!]   
     \caption{The proposed co-teaching framework} 
     \KwIn{model parameters $\theta_A$, $\theta_B$, learning rate $\eta$, number of epochs $n_T$, number of iterations $n_K$;} 
        \For{$T=1,2,...,T_{n_T}$} 
        { 
            \textbf{Shuffle} training set $\mathcal{D}$\;
            \For{$K=1,2,...,K_{n_K}$}
            {  
                \textbf{Fetch} a batch of training data $\Bar{\mathcal{D}}$\; 
                \textbf{Distributes} $\Bar{\mathcal{D}}$ equally to two sub-batches of training data $\Bar{\mathcal{D}}_A, \Bar{\mathcal{D}}_B$; \Comment{$\Bar{\mathcal{D}}_A, \Bar{\mathcal{D}}_B \subset \Bar{\mathcal{D}} $}\\
                \textbf{Obtain} learning protocol $(\Tilde{\mathcal{D}}_B, \mathcal{J}_B)$ from model A and $\Bar{\mathcal{D}}_B$\;  
                \textbf{Obtain} learning protocol $(\Tilde{\mathcal{D}}_A, \mathcal{J}_A)$ from model B and $\Bar{\mathcal{D}}_A$\;
                \textbf{Update} $\theta_A = \theta_A - \eta\nabla\mathcal{J}_A(\Tilde{\mathcal{D}}_A)$; \Comment{Update model A by $(\Tilde{\mathcal{D}}_A, \mathcal{J}_A)$.}\\
                \textbf{Update} $\theta_B = \theta_B - \eta\nabla\mathcal{J}_B(\Tilde{\mathcal{D}}_B)$; \Comment{Update model B by $(\Tilde{\mathcal{D}}_B, \mathcal{J}_B)$.}\\
            }
        } 
    \KwOut{$\theta_A$, $\theta_B$.} 
\end{algorithm*}

\subsection{Teaching Strategies}\label{teachstra}
We consider the following three strategies that cover teaching with dynamic loss functions and teaching with data curriculum.

\paragraph{Teaching with Dynamic Margins:} The strategy fixes $\Bar{\mathcal{D}}_A$ and $\Bar{\mathcal{D}}_B$ as $\Tilde{\mathcal{D}}_A$ and $\Tilde{\mathcal{D}}_B$ respectively, and dynamically creates loss functions as the learning protocols. Without loss of generality, the training data $\mathcal{D}$ can be re-organized in a form of $\{( c_i, r_i^+, r_i^-) \}_{i=1}^{N'}$, where $r_i^+$ and $r_i^-$ refer to a positive response candidate and a negative response candidate regarding to $c_i$ respectively. Suppose that $\Bar{\mathcal{D}}_A=\{(c_{A,i}, r_{A,i}^+, r_{A,i}^-)\}_{i=1}^{N_A}$ and $\Bar{\mathcal{D}}_B=\{(c_{B,i}, r_{B,i}^+, r_{B,i}^-)\}_{i=1}^{N_B}$, then model A evaluates each $(c_{B,i}, r_{B,i}^+, r_{B,i}^-) \in \Bar{\mathcal{D}}_B$ with matching scores $s_A(c_{B,i}, r_{B,i}^+)$ and $s_A(c_{B,i}, r_{B,i}^-)$, and form a margin for model B as
\begin{equation}\label{marginB} 
\fontsize{8.5pt}{10pt} 
\begin{aligned} 
    \Delta_{B,i}=\max\Big(0, \lambda \big(s_A(c_{B,i}, r_{B,i}^+) - s_A(c_{B,i}, r_{B,i}^-)\big)\Big),
\end{aligned}
\end{equation}
where $\lambda$ is a hyper-parameter. Similarly, $\forall (c_{A,i}, r_{A,i}^+, r_{A,i}^-) \in \Bar{\mathcal{D}}_A$, the margin provided by model B for model A can be formulated as 
\begin{equation}\label{marginA}  
\fontsize{8.5pt}{10pt} 
\begin{aligned} 
    \Delta_{A,i}=\max\Big(0, \lambda \big(s_B(c_{A,i}, r_{A,i}^+) - s_B(c_{A,i}, r_{A,i}^-)\big)\Big),
\end{aligned} 
\end{equation}
where $s_B(c_{A,i}, r_{A,i}^+)$ and $s_B(c_{A,i}, r_{A,i}^-)$ are matching scores calculated with model B. Loss functions $\mathcal{J}_A$ and $\mathcal{J}_B$ are then defined as
\begin{equation}\fontsize{10.2pt}{10pt} 
  \begin{aligned}   
    \mathcal{J}_A = \sum_{i=1}^{N_A} \max\{0, \Delta_{A,i} &  -s_A(c_{A,i}, r_{A,i}^+) \\ 
        & + s_A(c_{A,i}, r_{A,i}^-)\}, 
    \end{aligned}  
\end{equation}
\begin{equation}\fontsize{10.2pt}{10pt} 
\begin{aligned} 
    \mathcal{J}_B =  \sum_{i=1}^{N_B} \max\{0, \Delta_{B,i} & -s_B(c_{B,i}, r_{B,i}^+) \\
    & +s_B(c_{B,i}, r_{B,i}^-)\}.
\end{aligned}
\end{equation}
Intuitively, one model may assign a small margin to a negative example if it identifies the example as a false negative. Then, its peer model will pay less attention to such an example in its optimization. This is how the two peer models help each other combat with noise under the strategy of teaching with dynamic margins. 

\paragraph{Teaching with Dynamic Instance Weighting:} Similar to the first strategy, this strategy also defines the learning protocols with dynamic loss functions. The difference is that this strategy penalizes low-quality negative training examples with weights. Formally, let us represent $\Bar{\mathcal{D}}_B$ as $\{(y_{B,i}, c_{B,i}, r_{B,i})\}_{i=1}^{N'_B}$, then $\forall (y_{B,i}, c_{B,i}, r_{B,i}) \in \Bar{\mathcal{D}}_B$, its weight from model A is defined as
\begin{equation}
    w_{B,i}=
    \left\{
             \begin{array}{lr}
             1 & y_{B,i}=1 \\
             1 - s_A(c_{B, i}, r_{B, i}) & y_{B, i} = 0 
             \end{array}
    \right.
\end{equation}
Similarly, $\forall (y_{A,i}, c_{A,i}, r_{A,i}) \in \Bar{\mathcal{D}}_A$, model B assign a weight as
\begin{equation}
    w_{A, i}=
    \left\{
             \begin{array}{lr}
             1 & y_{A,i}=1 \\
             1 - s_B(c_{A,i}, r_{A,i}) & y_{A,i} = 0   
             \end{array}
    \right.
\end{equation}

Then, loss functions $\mathcal{J_A}$ and $\mathcal{J_B}$ can be formulated as
\begin{align}
   \mathcal{J}_A=\sum_{i=1}^{N'_A} w_{A,i} L(y_{A,i}, s_A(c_{A,i}, r_{A,i})),  \\
   \mathcal{J}_B=\sum_{i=1}^{N'_B} w_{B,i} L(y_{B,i}, s_B(c_{B,i}, r_{B,i})), 
\end{align}
where $L(\cdot,\cdot)$ is defined by cross entropy:
\begin{equation}
- y \log(s(c, r)) +(1-y) \log(1-s(c, r)).
\end{equation}

In this strategy, negative examples that are identified as false negatives by one model will obtain small weights from the model, and thus be less important than other examples in the learning process of the other model. 

\paragraph{Teaching with Dynamic Data Curriculum:} In the first two strategies, knowledge is transferred mutually through ``soft labels'' defined by the peer matching models. In this strategy, we directly transfer data to each model.  During learning, $\mathcal{J}_A$ and $\mathcal{J}_B$ are fixed as cross entropy, and the learning protocols vary by $\Tilde{\mathcal{D}}_A$ and $\Tilde{\mathcal{D}}_B$. Inspired by \newcite{BoHanNIPS2018}, we construct $\Tilde{\mathcal{D}}_A$ and $\Tilde{\mathcal{D}}_B$ with small-loss instances. These instances are far from decision boundaries of the two models, and thus are more likely to be true positives and true negatives.  Formally, $\Tilde{\mathcal{D}}_A$ and $\Tilde{\mathcal{D}}_B$ are defined as
\begin{equation} \normalsize 
\fontsize{10.7pt}{10pt} 
\label{datafilter}
    \begin{aligned} 
    \Tilde{\mathcal{D}}_B = argmin_{\left|\Tilde{\mathcal{D}}_B\right|=\delta\left|\Bar{\mathcal{D}}_B\right|, \Tilde{\mathcal{D}}_B \subset \Bar{\mathcal{D}}_B} \mathcal{J}_A (\Tilde{\mathcal{D}}_B), \\
    \Tilde{\mathcal{D}}_A = argmin_{\left|\Tilde{\mathcal{D}}_A\right|=\delta\left|\Bar{\mathcal{D}}_A\right|, \Tilde{\mathcal{D}}_A \subset \Bar{\mathcal{D}}_A} \mathcal{J}_B(\Tilde{\mathcal{D}}_A),
\end{aligned}
\end{equation}
where $|\cdot|$ measures the size of a set,  $\mathcal{J}_A (\Tilde{\mathcal{D}}_B)$ and $\mathcal{J}_B(\Tilde{\mathcal{D}}_A)$ stand for accumulation of loss on the corresponding data sets, and $\delta$ is a hyper-parameter. Note that we do not shrink $\delta$ as in \newcite{BoHanNIPS2018}, since fixing $\delta$ as a constant yields a simple yet effective learning program, as will be seen in our experiments. 

\section{Experiments}
We test our learning schemes on two public data sets with human annotated test examples.

\subsection{Experimental Setup}
The first data set we use is Douban Conversation Corpus (Douban) \cite{wu2017sequential} which is a multi-turn Chinese conversation data set crawled from Douban group\footnote{\url{https://www.douban.com/group}}. The data set consists of $1$ million context-response pairs for training, $50$ thousand pairs for validation, and $6,670$ pairs for test. In the training set and the validation set, the last turn of each conversation is regarded as a positive response and negative responses are randomly sampled. The ratio of the positive and the negative is $1$:$1$ in training and validation. In the test set, each context has $10$ response candidates retrieved from an index whose appropriateness regarding to the context is judged by human annotators. The average number of positive responses per context is $1.18$. Following \newcite{wu2017sequential}, we employ R$_{10}$@1, R$_{10}$@2, R$_{10}$@5, mean average precision (MAP), mean reciprocal rank (MRR), and precision at position 1~(P@1) as evaluation metrics.

In addition to the Douban data, we also choose E-commerce Dialogue Corpus (ECD) \cite{coling2018dua} as an experimental data set. The data consists of real-world conversations between customers and customer service staff in Taobao\footnote{\url{https://www.taobao.com}}, which is the largest e-commerce platform in China. There are $1$ million context-response pairs in the training set, and $10$ thousand pairs in both the validation set and the test set.  Each context in the training set and the validation set corresponds to one positive response candidate and one negative response candidate, while in the test set, the number of response candidates per context is $10$ with only one of them positive. In the released data, human responses are treated as positive responses, and negative ones are automatically collected by ranking the response corpus based on conversation history augmented messages using Apache Lucene\footnote{\url{http://lucene.apache.org/}}. Thus, we recruit $3$ active users of Taobao as human annotators, and ask them to judge each context-response pair in the test data (i.e., in total $10$ thousand pairs are judged). If a response can naturally reply to a message given the conversation history before it, then the context-response pair is labeled as $1$, otherwise, it is  labeled as $0$. Each pair receives three labels and the majority is taken as the final decision. On average, each context has $2.5$ response candidates labeled as positive. There are only $33$ contexts with all responses labeled as positive or negative, and we remove them from test. Fleiss' kappa \cite{fleiss1971measuring} of the labeling is $0.64$, indicating substantial agreement among the annotators. We employ the same metrics as in Douban for evaluation. 

Note that we do not choose the Ubuntu Dialogue Corpus \cite{lowe2015ubuntu} for experiments, because (1) the test set of the Ubuntu data is constructed by randomly sampling; and (2) conversations in the Ubuntu data are in a casual style and too technical, and thus it is very difficult for us to find qualified human annotators to label the data.

 \begin{table*}[t!]
    \centering
    \resizebox{\textwidth}{!}{
    \begin{tabular}{|l|c|c|c|c|c|c|c|c|c|c|c|c|}
      \hline
      &   \multicolumn{6}{c|}{\textbf{Douban}}    &        \multicolumn{6}{c|}{\textbf{ECD}}        \\ \cline{2-13}
      & MAP & MRR & P@1 & R$_{10}$@1 & R$_{10}$@2  & R$_{10}$@5  &  MAP & MRR & P@1 &R$_{10}$@1 & R$_{10}$@2  & R$_{10}$@5 \\ \hline \hline
      SMN~\cite{wu2017sequential} & 0.529 & 0.569 & 0.397 & 0.233 & 0.396 & 0.724 & - & - & - & - & - & - \\
      SMN-Pre-training         & 0.527 & 0.570 & 0.396 & 0.236 & 0.392 & 0.734 & 0.662 & 0.742 & 0.598 & 0.302 & 0.464 & 0.757 \\ \hline
      SMN-Margin               & \textbf{0.559}$^*$ &	\textbf{0.601}$^*$ & \textbf{0.424}$^*$ & \textbf{0.260}$^*$ & \textbf{0.426}$^*$ & \textbf{0.764}$^*$ & 0.674 & 0.750 & 0.615 & 0.318 & 0.481 & 0.765  \\ 
      SMN-Weighting            & 0.550$^*$ & 0.593$^*$ & 0.414 & 0.253 & 0.413 & 0.762$^*$ & 0.666 & 0.745 & 0.601 & 0.311 & 0.475 & 0.775 \\
      SMN-Curriculum           & 0.548 & 0.594$^*$ & 0.418$^*$ & 0.254$^*$ & 0.411 & 0.763$^*$ & \textbf{0.678} & \textbf{0.762}$^*$ & \textbf{0.622}$^*$ & \textbf{0.323}$^*$ & \textbf{0.487}$^*$ & \textbf{0.778}$^*$ \\ \hline \hline
      DAM~\cite{zhou2018multi}    & 0.550 & 0.601 & 0.427 & 0.254 & 0.410 & 0.757 &  - & - & - & - & - & -  \\
      DAM-Pre-training         & 0.552 & 0.605 & 0.426 & 0.258 & 0.408 & 0.766 & 0.685 & 0.756 & 0.621 & 0.325 & 0.491 & 0.772 \\ \hline
      DAM-Margin               & \textbf{0.583}$^*$ & 0.628$^*$ &	0.451$^*$ &	\textbf{0.276}$^*$ & \textbf{0.454$^*$} & 0.806$^*$ & 0.692 & \textbf{0.777$^*$}	& 0.652$^*$	& 0.337	& 0.506	& 0.778  \\ 
      DAM-Weighting            & 0.579$^*$ & \textbf{0.629}$^*$ & \textbf{0.453}$^*$ & 0.272 & \textbf{0.454}$^*$ & \textbf{0.809}$^*$ & 0.695 & 0.775 & 0.651$^*$ & 0.343 & 0.497 & \textbf{0.789} \\
      DAM-Curriculum           & 0.580$^*$ & 0.623$^*$ & 0.442 & 0.269 & 0.459$^*$ & 0.804$^*$ & \textbf{0.696} & \textbf{0.777}$^*$ & \textbf{0.653}$^*$ & \textbf{0.345}$^*$ & \textbf{0.506} & 0.781 \\
      \hline
    \end{tabular}
    }
    \caption{Evaluation results on the two data sets. Numbers marked with $*$ mean that the improvement is statistically significant compared with the best baseline (t-test with $p$-value $<0.05$). Numbers in bold indicate the best strategies for the corresponding models on specific metrics.}  
    \label{exp:main-results}
 \end{table*}

\subsection{Matching Models}
We select the following two models that achieve superior performance on benchmarks to test our learning approach. 

\textbf{SMN:}~\cite{wu2017sequential} first lets each utterance in a context interact with a response, and forms a matching vector for the pair through CNNs. Matching vectors of all the pairs are then aggregated with an RNN as a matching score.

\textbf{DAM:}~\cite{zhou2018multi} performs matching under a representation-matching-aggregation framework, and represents a context and a response with stacked self-attention and cross-attention.   

Both models are implemented with TensorFlow according to the details in \newcite{wu2017sequential} and \newcite{zhou2018multi}. To implement co-teaching, we pre-train the two models using the training sets of Douban and ECD, and tune the models with the validation sets of the two data. Each pre-trained model is used to initialize both model A and model B. After co-teaching, the one in A and B that performs better on the validation sets is picked for comparison. We denote models learned with the teaching strategies in Section \ref{teachstra} as Model-Margin, Model-Weighting, and Model-Curriculum respectively, where ``Model'' refers to either SMN or DAM. These models are compared with the pre-trained model denoted as Model-Pre-training, and those reported in \newcite{wu2017sequential,zhou2018multi,coling2018dua}.

\subsection{Implementation Details}
We limit the maximum number of utterances in each context as $10$ and the maximum number of words in each utterance and response as $50$ for computational efficiency. Truncation or zero-padding are applied when necessary. Word embedding is pre-trained with Word2Vec~\cite{mikolov2013distributed} on the training sets of Douban and ECD, and the dimension of word vectors is $200$. The co-teaching framework is implemented with TensorFlow. In co-teaching, learning rates (i.e., $\eta$ in Algorithm 1) in dynamic margins, dynamic instance weighting, and dynamic data curriculum are set as $0.001$, $0.0001$, and $0.0001$ respectively. We choose $200$ in co-teaching with SMN and $50$ in co-teaching with DAM as the size of mini-batches. Optimization is conducted using stochastic gradient descent with Adam algorithm~\cite{kingma2014adam}. In teaching with dynamic margins, we vary $\lambda$ in $\{1, \frac{1}{2}, \frac{1}{3}, \frac{1}{5}, \frac{1}{10}, \frac{1}{15}, \frac{1}{20}\}$, and choose $\frac{1}{10}$ for SMN on Douban, $\frac{1}{2}$ for SMN on ECD,  $\frac{1}{3}$ for DAM on Douban, and $\frac{1}{2}$ for DAM on ECD. In teaching with dynamic data curriculum, we select $\delta$ in $\{0.1, 0.2, ..., 0.9, 1.0\}$, and find that $0.9$ is the best choice for both models on both data sets. 

\subsection{Evaluation Results}
Table \ref{exp:main-results} reports evaluation results of co-teaching with the three teaching strategies on the two data sets. We can see that all teaching strategies can improve the original models on both data sets, and improvement from the best strategy is statistically significant (t-test with $p$-value $<0.05)$ on most metrics. On Douban, the best strategy for SMN is teaching with dynamic margins, and it is comparable with teaching with dynamic instance weighting for DAM,  while on ECD, for both SMN and DAM, the best strategy is teaching with dynamic data curriculum. The difference may stem from the  nature of training sets of the two data. The training set of Douban is built from random sampling, while the training set of ECD is constructed through response retrieval that may contain more false negatives. Thus, in training, Douban could be cleaner than ECD, making ``hard data filtering'' more effective than ``soft labeling'' on ECD. It is worth noting that on ECD, there are significant gaps between the results of SMN (pre-trained) reported in Table \ref{exp:main-results} and those reported in \newcite{coling2018dua}, since SMN in this paper is evaluated on the human-judged test set while SMN in \newcite{coling2018dua} is evaluated on the automatically constructed test set that is homogeneous with the training set. This somehow indicates the gap between training  and test in real applications for the existing research on response selection, and thus demonstrates the merits of this work.

\subsection{Discussions}
In addition to efficacy of co-teaching as a learning approach, we are also curious about \textbf{Q1:} if model A and model B can ``co-evolve''  when they are initialized with one network; \textbf{Q2:} if co-teaching is still effective when model A and model B are initialized with different networks; and \textbf{Q3:} if the teaching strategies are sensitive to the hyper-parameters (i.e., $\lambda$ in Equations (\ref{marginB})-(\ref{marginA}) and $\delta$ in Equation (\ref{datafilter})).

\begin{figure*}[t!]
  \centering
  \subfigure[Dynamic data curriculum] { \label{fig:trend_cl}
    \includegraphics[width=0.65\columnwidth]{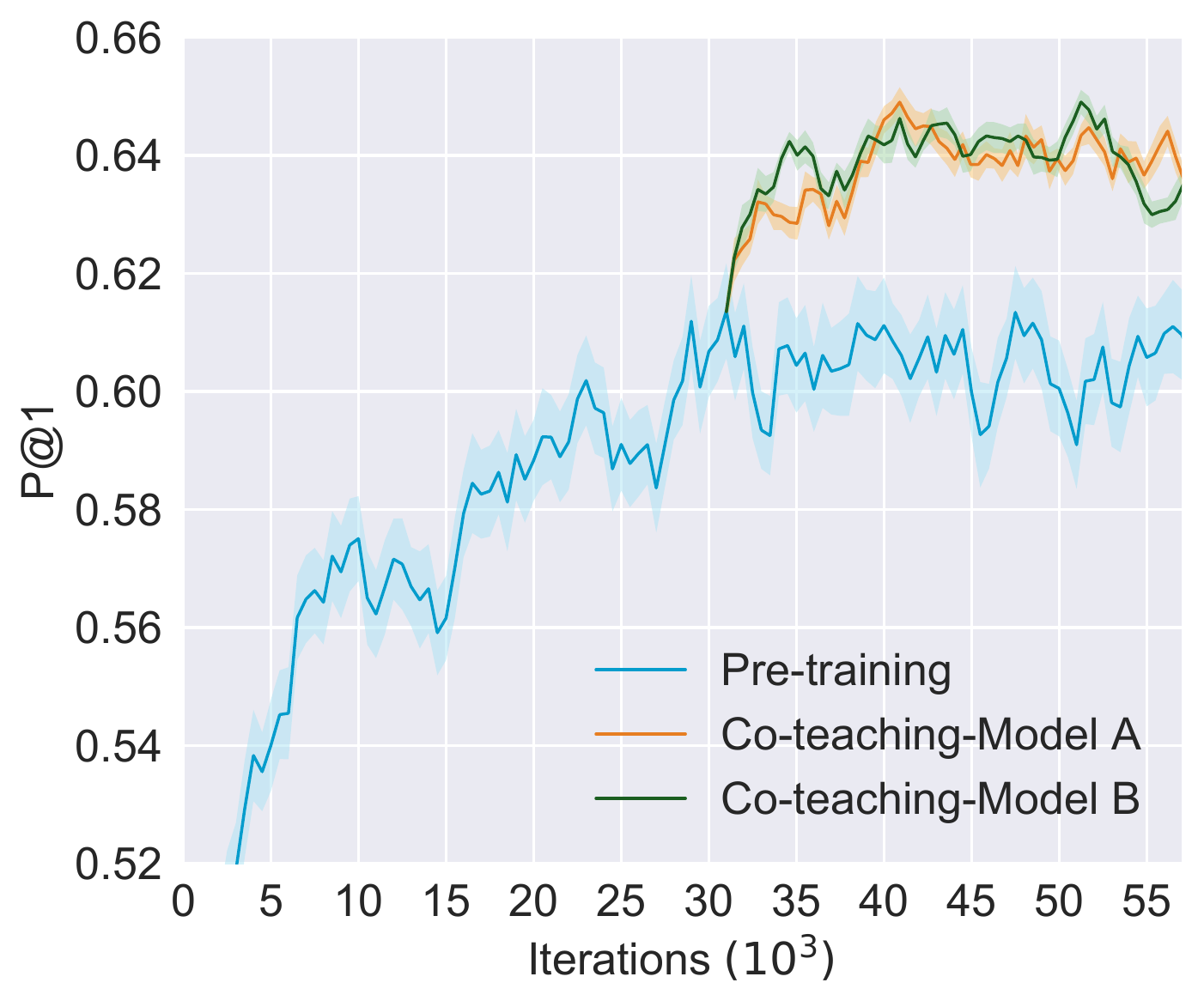}
  } 
  \subfigure[Dynamic instance weighting] { \label{fig:trend_iw}
    \includegraphics[width=0.65\columnwidth]{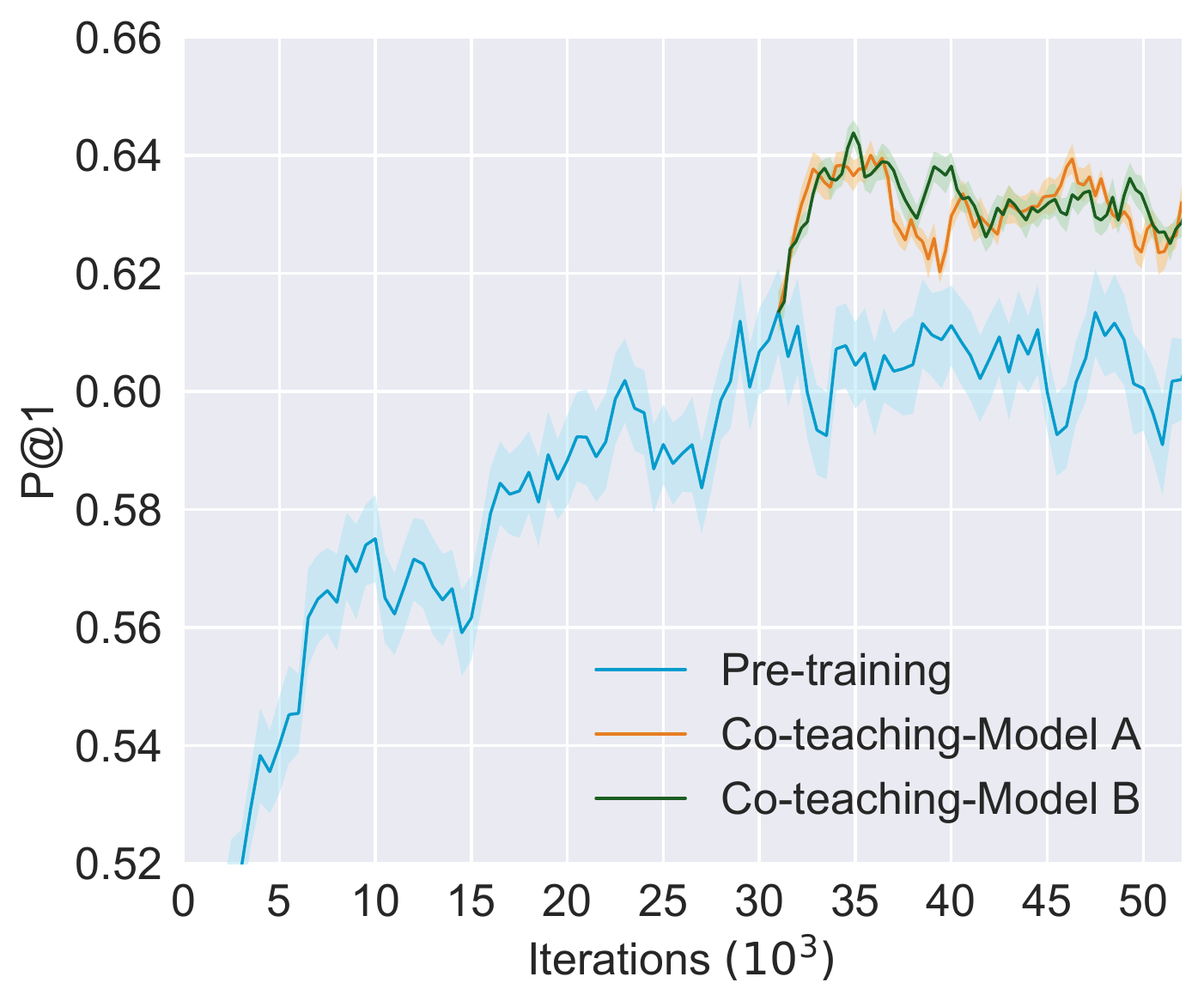}
  }
  \subfigure[Dynamic margins] { \label{fig:trend_margin}
    \includegraphics[width=0.65\columnwidth]{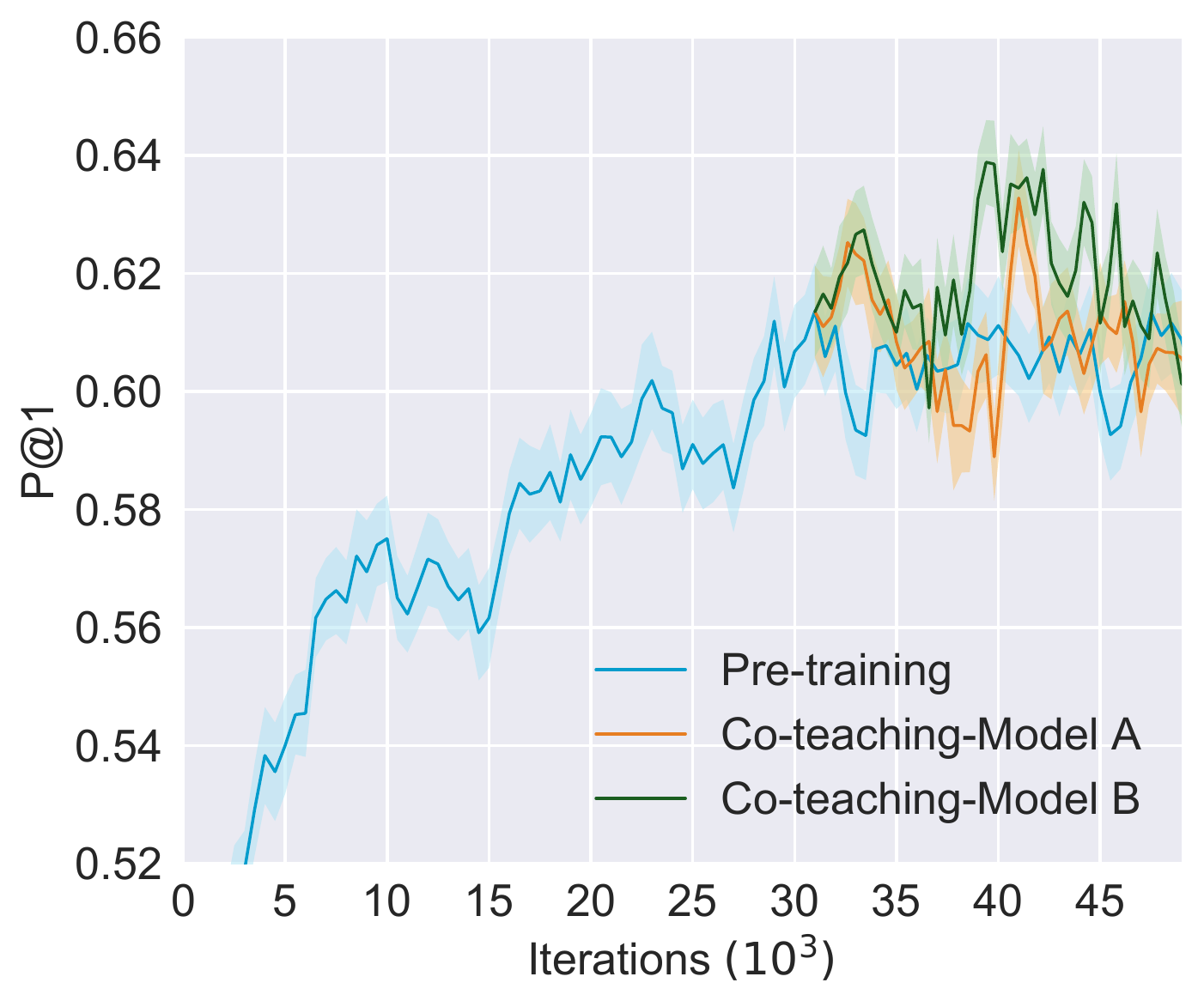}
  }  
  \caption{Test P@1 of DAM with the three teaching strategies on ECD. All curves are smoothed by exponential moving average\protect\footnotemark \ for beauty.}
  \label{fig:P@1_curve}
\end{figure*}


 \begin{table*}[t!]
    \centering
    \resizebox{\textwidth}{!}{
      \begin{tabular}{|l|c|c|c|c|c|c|c|c|c|c|c|c|}
      \hline
      &   \multicolumn{6}{c|}{\textbf{Douban (Margin)}}    &        \multicolumn{6}{c|}{\textbf{ECD (Curriculum)} }        \\ \cline{2-13} 
      & MAP & MRR & P@1 & R$_{10}$@1 & R$_{10}$@2  & R$_{10}$@5  &  MAP & MRR & P@1 &R$_{10}$@1 & R$_{10}$@2  & R$_{10}$@5 \\ \hline \hline
      SMN-Pre-training  & 0.527 & 0.570 & 0.396 & 0.236 & 0.392 & 0.734 
                        & 0.662 & 0.742 & 0.598 & 0.302 & 0.464 & 0.757 \\ 
      SMN-Co-teaching & 0.558 & 0.602 & 0.420 & 0.255 & 0.431 & 0.787 & 0.674 & 0.765 & 0.626 & 0.322 & 0.485 & 0.779                         \\ \hline 
      DAM-Pre-training & 0.552 & 0.605 & 0.426 & 0.258 & 0.408 & 0.766 & 0.685 & 0.756 & 0.621 & 0.325 & 0.491 & 0.772
                             \\ 
      DAM-Co-teaching & 0.570 & 0.617 & 0.438 & 0.270 & 0.455 & 0.781  & 0.696 & 0.775 & 0.652 & 0.341 & 0.499 & 0.784  \\ 
      \hline
    \end{tabular}
    }
    \caption{Evaluation results of co-teaching initialized with different networks.}
    \label{exp:strong_and_weak-results}
 \end{table*}

\paragraph{Answer to Q1:} Figure \ref{fig:P@1_curve} shows P@1 of DAM vs. number of iterations on the test set of ECD under the three teaching strategies. Co-teaching with any of the three strategies can improve both  the performance of model A and the performance of model B after pre-training, and the peer models move with almost the same pace. The results verified our claim that ``by learning from each other, the peer models can get improved together''.  Curves of dynamic margins oscillate more fiercely than others, indicating that optimization with dynamic margins is more difficult than optimization with the other two strategies. 

\paragraph{Answer to Q2:} as a case study of co-teaching with two networks in different capabilities, we initialize model A and model B with DAM and SMN respectively, and select teaching with dynamic margins for Douban and teaching with dynamic data curriculum for ECD (i.e., the best strategies for the two data sets when co-teaching is initialized with one network). Table \ref{exp:strong_and_weak-results} shows comparison between models before/after co-teaching. We find that co-teaching is still effective when starting from two networks, as both SMN and DAM get improved on the two data sets. Despite the improvement, it is still better to learn the two networks one by one,  as co-teaching with two  networks cannot bring more improvement than co-teaching with one network, and the performance of the stronger one between the two networks could also drop (e.g., DAM on Douban). We guess this is because the stronger model cannot be well taught by the weaker model, especially in teaching via “soft labels”, and as a result, it is not able to transfer more knowledge to the weaker one as well.

\footnotetext{\url{https://en.wikipedia.org/wiki/Moving_average\#Exponential\_moving\_average}}

\begin{figure}[t!]
  \centering
    \subfigure[\scriptsize Dynamic margins on Douban] { \label{fig:DAM_MARGIN_LAMBDA}
    \includegraphics[width=0.485\columnwidth]{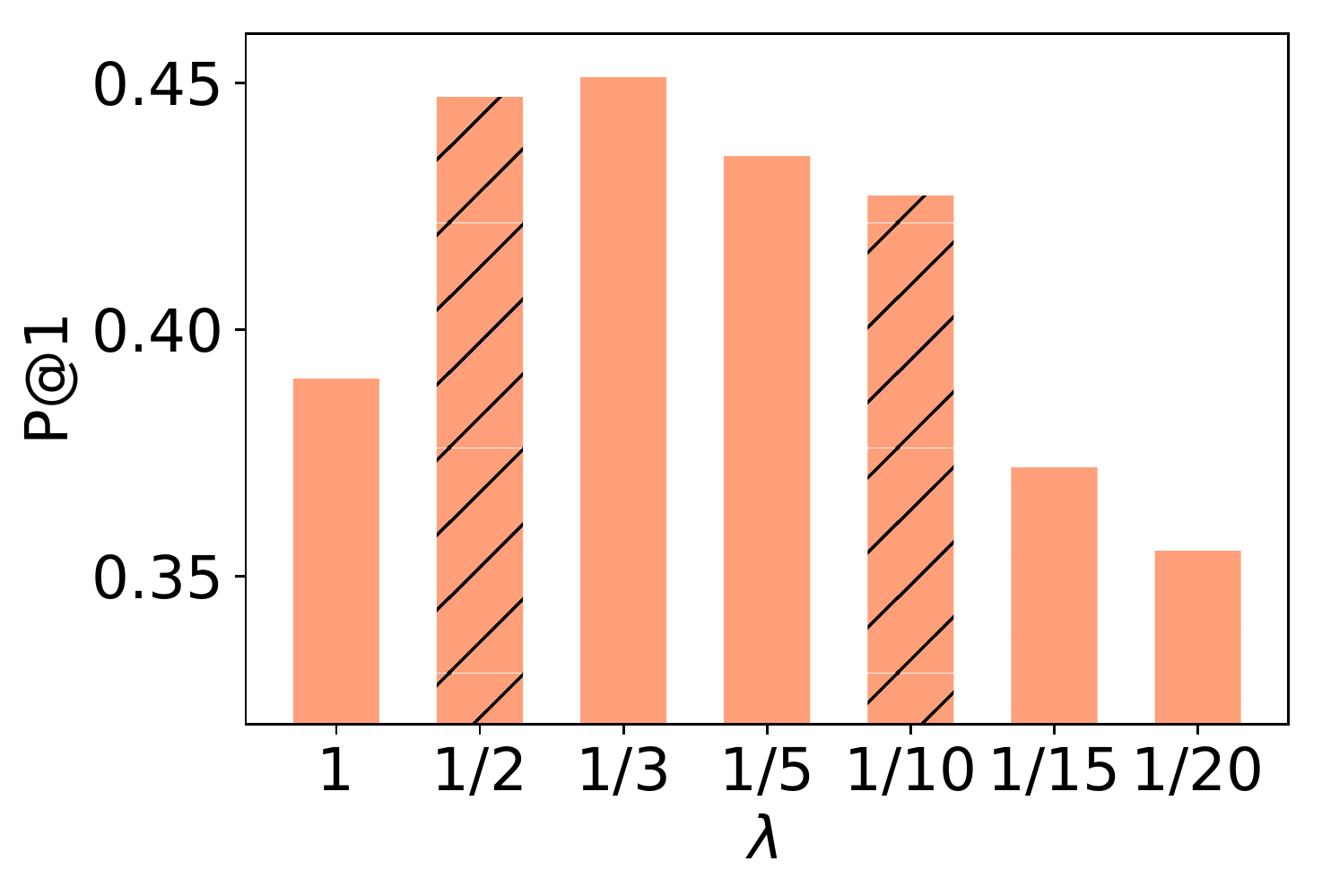} } \hspace{-4mm}
    \subfigure[\scriptsize Data curriculum on ECD] { \label{fig:DAM_CURRI_DELTA}
    \includegraphics[width=0.485\columnwidth]{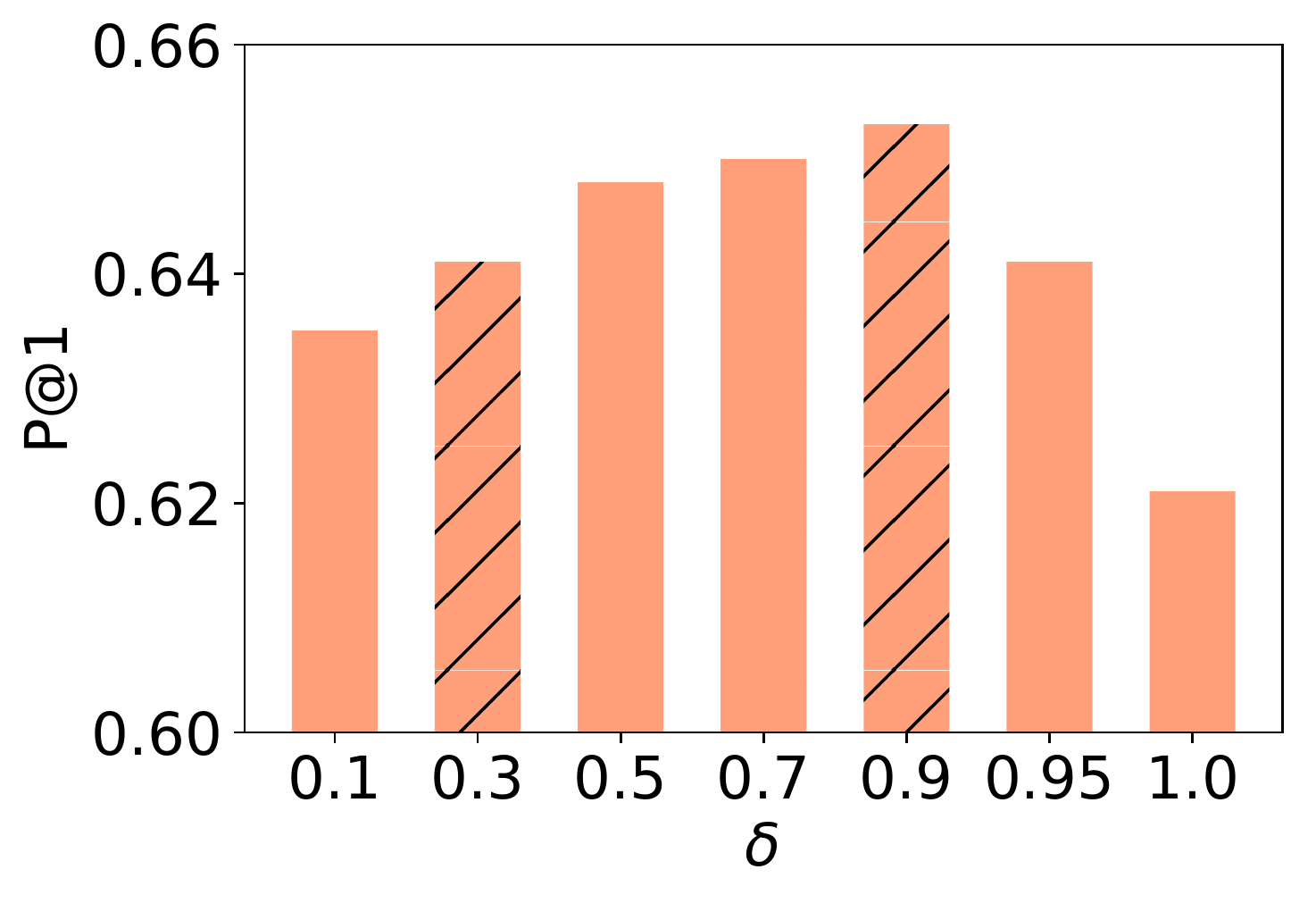} } 
    \caption{Effects of $\lambda$ and $\delta$ to co-teaching. Experiments are conducted with DAM on the two data sets.}
    \label{fig:delta}
\end{figure}

\paragraph{Answer to Q3:} finally, we check the effect of hyper-parameters to co-teaching. Figure \ref{fig:DAM_MARGIN_LAMBDA} illustrates how the performance of DAM varies under different $\lambda$s in teaching with dynamic margins on Douban. We can see that both small $\lambda$s and large $\lambda$s will cause performance drop. This is because small $\lambda$s will reduce the effect of margins, making clean examples and noisy examples indifferent in learning, while with large $\lambda$s, some errors from the ``soft labels'' might be magnified, and thus hurt the performance of the learning approach.  Figure \ref{fig:DAM_CURRI_DELTA} shows the performance of DAM under different $\delta$s in teaching with dynamic data curriculum on ECD.  Similarly, DAM gets worse when $\delta$ becomes small or large, since a smaller $\delta$ means fewer data will be involved in training, while a larger $\delta$ brings more risks to introducing noise into training. Thus, we conclude that the teaching strategies are sensitive to the choice of hyper-parameters.

\section{Related Work}
So far, methods used to build an open domain dialogue system can be divided into two categories. The first category utilize an encoder-decoder framework to learn response generation models. Since
the basic sequence-to-sequence models \cite{vinyals2015neural,shangL2015neural,tao2018get} tend to generate generic responses, extensions have been made to incorporate external knowledge into generation~\cite{mou2016sequence,xing2017topic}, and to generate responses with specific personas or emotions \cite{li2016persona,zhang2018personalizing,zhou2017emotional}.
The second category design a discriminative model to measure the matching degree between a human input and a response candidate for response selection. At the beginning, research along this line assumes that the human input is a single message \cite{lu2013deep,wang2013dataset,hu2014convolutional,wang2015syntax}. Recently, researchers begin to make use of conversation history in matching. Representative methods include the dual LSTM model~\cite{lowe2015ubuntu}, the deep learning to respond architecture~\cite{rui2018learning}, the multi-view matching model~\cite{zhou2016multi}, the sequential matching network~\cite{wu2017sequential,wu2017sequentialframe}, the deep attention matching network~\cite{zhou2018multi}, and the multi-representation fusion network~\cite{tao2019multi}. 

Our work belongs to the second group. Rather than crafting a new model, we are interested in how to learn the existing models with a better approach. Probably the most related work is the weakly supervised learning approach proposed in \newcite{wu2018learning}. However, there is stark difference between our approach and the weak supervision approach: (1) weak supervision employs a static generative model to teach a discriminative model, while co-teaching dynamically lets two discriminative models teach each other and evolve together; (2) weak supervision needs pre-training a generative model with extra resources and pre-building an index for training data construction, while co-teaching does not have such request; and (3) in terms of multi-turn response selection, weak supervision is only tested on the Douban data with SMN and the multi-view matching model, while co-teaching is proven effective on both the Douban data and the E-commerce data with SMN and DAM which achieves state-of-the-art performance on benchmarks. Moreover, improvement to SMN on the Douban data from co-teaching is bigger than that from weak supervision, when the ratio of the positive and the negative is 1:1 in training\footnote{Our results are $0.559$ (MAP), $0.601$ (MRR), and $0.424$ (P@1), while results reported in \cite{wu2018learning} are $0.542$ (MAP), $0.588$ (MRR), and $0.408$ (P@1).}.

Our work, in a broad sense, belongs to the effort on learning with noisy data. Previous studies including curriculum learning (CL)~\cite{bengio2009curriculum} and self-paced learning (SPL)~\cite{jiang2014self,jiang2015self} tackle the problem with heuristics, such as ordering data from easy instances to hard ones~\cite{spitkovsky2010baby,tsvetkov2016learning} and retaining training instances whose losses are smaller than a threshold ~\cite{jiang2015self}. Recently, \newcite{fan2018learning} propose a deep reinforcement learning framework in which a simple deep neural network is used to adaptively select and filter important data instances from the training data. \newcite{jiang2017mentornet} propose a MentorNet which learns a data-driven curriculum with a Student-Net to mitigate overfitting on corrupted labels. In parallel to curriculum learning,  several studies explore sample weighting schemes where training samples are re-weighted according to their label-quality~\cite{wang2017robust,dehghani2018fidelity,wu2018learning}.  Instead of considering data quality, \newcite{wu2018NIPSL2T-DLF} employ a parametric model to dynamically create appropriate loss functions.

The learning approach in this work is mainly inspired by the work of \newcite{BoHanNIPS2018} for handling extremely noisy labels. However, with substantial extensions, our work is far beyond that work. First, we generalize the concept of ``co-teaching'' to a framework, and now the method in \newcite{BoHanNIPS2018} becomes a special case of the framework. Second, \newcite{BoHanNIPS2018} only exploits data curriculum, while in addition to data curriculum, we also propose two new strategies for teaching with dynamic loss functions as special cases of the framework. Third, unlike \newcite{BoHanNIPS2018} who only use one network to initialize the peer models in co-teaching,  we studied co-teaching with both one network and two different networks.  Finally, \newcite{BoHanNIPS2018} verified that the special co-teaching method is effective in some computer vision tasks, while we demonstrate that the co-teaching framework is generally useful for building retrieval-based dialogue systems.

\section{Conclusions}
We propose learning a matching model for response selection under a general co-teaching framework with three specific teaching strategies. The learning approach lets two matching models teach each other and evolve together. Empirical studies on two public data sets show that the proposed approach can generally improve the performance of existing matching models. 

\section*{Acknowledgement}
We would like to thank the anonymous reviewers for their constructive comments. This work was supported by the National Key Research and Development Program of China (No. 2017YFC0804001), the National Science Foundation of China (NSFC Nos. 61672058 and 61876196). 

\bibliography{acl2019}
\bibliographystyle{acl_natbib}

\clearpage

\end{document}